\def\year{2022}\relax
\documentclass[letterpaper]{article} 
\usepackage{aaai22}  
\usepackage{times}  
\usepackage{helvet}  
\usepackage{courier}  
\usepackage[hyphens]{url}  
\usepackage{graphicx} 
\usepackage{natbib}  
\usepackage{caption} 
\DeclareCaptionStyle{ruled}{labelfont=normalfont,labelsep=colon,strut=off} 
\usepackage{algorithm}
\usepackage{algorithmic}
\usepackage{newfloat}
\usepackage{listings}
\floatstyle{ruled}
\newfloat{listing}{tb}{lst}{}
\floatname{listing}{Listing}
\usepackage{multicolrule}

\usepackage{graphicx}
\usepackage{amsmath,amsfonts,amssymb}
\usepackage{mathtools}
\usepackage{caption}
\usepackage{lipsum}
\usepackage{booktabs}
\usepackage{multirow}
\usepackage{xcolor}
\colorlet{myPurple}{blue!60!red}
\definecolor{ao(english)}{rgb}{0.0, 0.5, 0.0}
\definecolor{cardinal}{rgb}{0.77, 0.12, 0.23}
\usepackage{color, colortbl}
\definecolor{Gray}{gray}{0.86}
\definecolor{LightGray}{gray}{0.94}
\definecolor{ao(english)}{rgb}{0.0, 0.5, 0.0}
\definecolor{cardinal}{rgb}{0.77, 0.12, 0.23}
\definecolor{cardinal}{rgb}{0.77, 0.12, 0.23}
\usepackage{url}
\usepackage{float}
\usepackage{tabularx}
\usepackage{appendix}
\usepackage{placeins}
\usepackage{enumitem}
\setlist[itemize]{leftmargin=*}
\makeatletter
\newcommand{\ssymbol}[1]{^{\@fnsymbol{#1}}}
\makeatother
\DeclareMathOperator*{\argmax}{argmax}
\title{Predicting Above-Sentence Discourse Structure using Distant Supervision from Topic Segmentation}
\author{
    Patrick Huber\equalcontrib,
    Linzi Xing\equalcontrib,
    Giuseppe Carenini
}
\affiliations{
    Department of Computer Science\\
    University of British Columbia\\
    Vancouver, BC, Canada, V6T 1Z4\\
}

\begin{document}

\maketitle

\begin{abstract}
RST-style discourse parsing plays a vital role in many NLP tasks, revealing the underlying semantic/pragmatic structure of potentially complex and diverse documents. Despite its importance, one of the most prevailing limitations in modern day discourse parsing is the lack of large-scale datasets. To overcome the data sparsity issue, distantly supervised approaches from tasks like sentiment analysis and summarization have been recently proposed. Here, we extend this line of research by exploiting distant supervision from topic segmentation, which can arguably provide a strong and oftentimes complementary signal for high-level discourse structures. Experiments on two human-annotated discourse treebanks confirm that our proposal generates accurate tree structures on sentence and paragraph level, consistently outperforming previous distantly supervised models on the sentence-to-document task and occasionally reaching even higher scores on the sentence-to-paragraph level.
\end{abstract}

\section{Introduction}
The Rhetorical Structure Theory (RST) \cite{mann1988rhetorical} is arguably one of the most popular frameworks to represent the 
discourse structure of complete documents. As such, RST has received considerable attention over the past decades, being leveraged to benefit important NLP downstream tasks, such as text classification \cite{ji-smith-2017-neural}, sentiment analysis \cite{bhatia-etal-2015-better, hogenboom2015using,  nejat-etal-2017-exploring, huber-carenini-2020-sentiment}, summarization \cite{Marcu1999, gerani-etal-2014-abstractive, xu-etal-2020-discourse, xiao-etal-2020-really} and argumentation analysis \cite{chakrabarty-etal-2019-ampersand}.

Compared to other competing discourse theories (e.g., PDTB \cite{prasad-etal-2008-penn}), RST postulates complete discourse trees based on clause-like Elementary Discourse Units (abbreviated: EDUs), which are sentence fragments on an intermediate granularity level between words and sentences. With EDUs acting as leaf nodes in RST-style discourse trees, constituents are formed by aggregating EDUs into sub-trees containing:
(1) a projective tree structure, (2) a nuclearity assignment for each internal node and (3) rhetorical relations between siblings. In this work, we focus on ``plain'' discourse tree structures and leave the exploration of nuclearity and relation classification for future work.

Due to the definition of complete discourse trees in the RST framework, tree structures become deeper as documents grow longer (as compared to the definition in PDTB, where the aggregation stops above sentence level). Furthermore, the defining factors for the tree aggregation on higher levels diverts considerably from the ones on lower levels (e.g., aggregating multiple paragraphs vs. combining EDUs)  \cite{jiang2021hierarchical}.
For example, suitable features for EDU level tree aggregations (i.e., low-levels) are mostly influenced by local syntactic and semantic signals, while the tree aggregation on paragraph level (i.e., high-levels) is likely to follow more global features, such as major topic shifts planned by the author for possibly complex communicative goals \cite{stede2011discourse}.

\begin{figure}[t]
    \setlength{\belowcaptionskip}{-15pt}
    \centering
    \includegraphics[width=.95\linewidth]{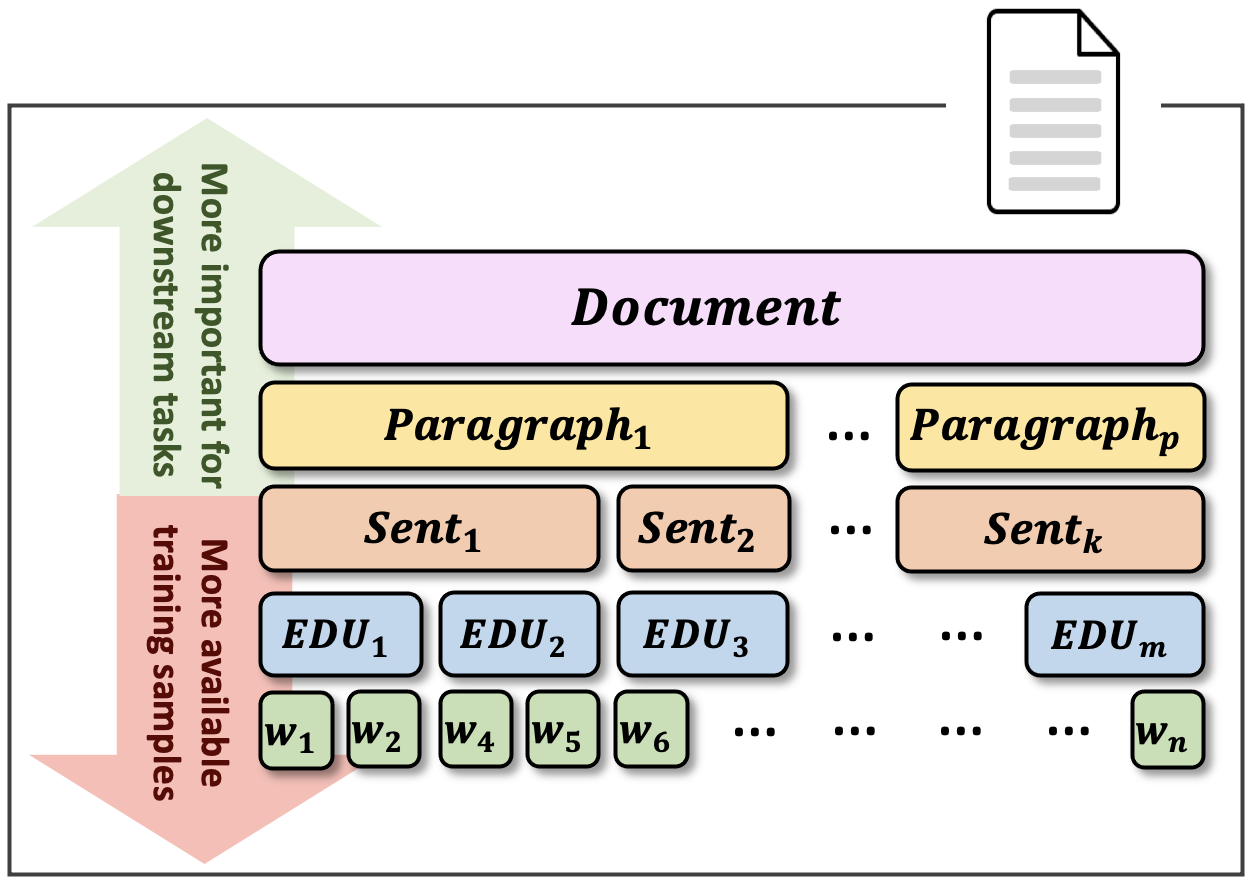}
    \caption{Textual levels in a well-structured document and their influence on downstream tasks and available gold-label data.}
    \label{fig:intro_example}
\end{figure}

Researchers working on RST-style discourse parsing take these considerations into account by either 
(1) proposing hard constraints to construct discourse trees on distinct textual levels by using varying feature sets \cite{ji-eisenstein-2014-representation, joty-etal-2015-codra} or  
(2) as done in recent work by \citet{wang-etal-2017-two} and \citet{guz-carenini-2020-coreference}, by proposing soft constraints, encoding sentence- and paragraph-breaks as input features. 
Furthermore, recent work on tree-level dependent RST-style discourse parsing mostly focuses on high-level tree structures \cite{jiang2021hierarchical}. 

Ideally, an RST-style discourse parser should achieve good performance on all levels shown in Figure~\ref{fig:intro_example}.
However, as argued in \citet{kobayashi2020top}, the performance on high-level constituency tree structures is especially important (green arrow in Figure~\ref{fig:intro_example}) when converting these constituency trees into dependency structures, as typically done for key downstream applications \cite{Marcu1999, marcu2000theory, ji-smith-2017-neural, shiv2019novel, huber-carenini-2020-sentiment, xiao2021predicting}.

Unfortunately, training samples for these critical high-level discourse structures are extremely limited (see red arrow in Figure \ref{fig:intro_example}). 
Not only does the largest available human-annotated treebank in English just contains 385 documents, but for each document, vastly more training samples are available for structures within sentences than for structures connecting sentences and paragraphs, since the number of nodes in a binary tree decreases exponentially from the leaves towards the root.

To tackle the data sparsity issue for discourse parsing, previous work has proposed to leverage distant supervision
from tasks with naturally annotated and abundantly available training data, such as sentiment analysis \cite{huber-carenini-2020-mega} and summarization \cite{xiao2021predicting}. While we believe that both these auxiliary tasks capture some structural information, they are plausibly even more aligned with other aspects of discourse, such as nuclearity for summarization and discourse relations (e.g., evidence, concession) for sentiment.
In contrast, this paper focuses exclusively on high-level tree structure generation, by exploiting signals from the auxiliary task of topic segmentation. Training on the naturally occurring and abundantly available topic segmentation annotations (sections/paragraphs), we believe that valuable information on major topic shifts, an effective signal indicative for high-level discourse tree structures, can be learned \cite{stede2011discourse}.

More specifically, we train the top-performing neural topic segmentation model proposed in \citet{xing-etal-2020-improving} and use the trained model to generate discourse structures, which we evaluate on two popular discourse parsing treebanks. Based on the sequential output of the topic segmentation model, we explore the generation of discourse trees using (1) a greedy top-down algorithm and (2) an optimal bottom-up CKY dynamic programming approach \cite{jurafsky2014speech}, predicting RST-style tree structures on/above sentence level. 

To better understand and properly compare the performance of our discourse tree generation algorithm with previously published
models, as well as a set of baselines, we evaluate all approaches on three partially overlapping discourse tree subsets from: sentence-to-paragraph (S-P), paragraph-to-document (P-D) and sentence-to-document (S-D). In our evaluation, we find that distant supervision from topic segmentation achieves promising results on the high-level tree structure generation task, consistently outperforming previous methods with distant supervision on sentence-to-document level and in some cases reaching superior performance compared to supervised models.

\section{Related Work}
\textbf{Rhetorial Structure Theory (RST) } \cite{mann1988rhetorical} is one of the main guiding theories for discourse parsing. As such, the RST framework proposes complete constituency discourse trees by first splitting a document into Elementary Discourse Units (EDUs) and subsequently aggregating them into larger (internal) sub-trees. The generated, projective tree structure (also called tree-span) is further augmented with a local importance score (called nuclearity), indicating a sub-tree as either a ``Nucleus'' (of primary significance) or a ``Satellite'' (of supplementary importance). 
Furthermore, rhetorical relations (e.g., evaluation, attribution and contrast) are assigned between adjacent sub-trees to represent the type of connection.  
In this paper, we follow the RST paradigm, generating above-sentence, ``plain'' constituency discourse trees (without nuclearity and relations) through distant supervision from topic segmentation.

Building on top of the RST discourse theory, \textbf{RST-style Discourse Parsing} aims to automatically infer discourse trees for new and unseen documents. 
With only a few small RST-style discourse treebanks available in English (e.g., RST-DT \cite{carlson2002rst}, Instruction-DT \cite{subba-di-eugenio-2009-effective} and GUM \cite{Zeldes2017}), most research in supervised discourse parsing has been focused on traditional machine learning approaches, such as DPLP \cite{ji-eisenstein-2014-representation}, CODRA \cite{joty-etal-2015-codra} and the Two-Stage parser \cite{wang-etal-2017-two}. More recently, a small number of neural solutions \cite{yu-etal-2018-transition, kobayashi2020top, guz-carenini-2020-coreference, guz-carenini-2020-towards} have also been proposed. However, due to the limited size of available treebanks, supervised models have been shown to not perform well when transferred across domains \cite{huber-carenini-2020-mega}.
As a result, a stream of unsupervised \cite{kobayashi-etal-2019-split, nishida-nakayama-2020-unsupervised, huber2021unsupervised} and distantly supervised \cite{huber-carenini-2019-predicting, huber-carenini-2020-mega, xiao2021predicting} discourse parsers have been proposed to address the data sparsity issue. Typically, distantly supervised approaches leverage supervision signals learned from tasks with abundant training data (e.g., sentiment analysis and extractive summarization) to infer discourse structures. Unsupervised approaches, on the other hand, mostly exploit the document hierarchy and separately aggregate tree structures on different levels based on recursively computed dissimilarity scores \cite{kobayashi-etal-2019-split}, syntactic knowledge \cite{nishida-nakayama-2020-unsupervised} or using an auto-encoder objective \cite{huber2021unsupervised}. In this paper, we propose another distantly supervised approach, computing discourse structures from the task of topic segmentation. 
We thereby draw inspiration from the findings in \citet{jiang2021hierarchical}, showing that a pre-trained topic segmenter can benefit supervised discourse parsers, especially on high-levels. However, instead of augmenting a supervised discourse parser with information obtained from a topic segmentation model, we explore the potential of directly inferring high-level discourse structures from the output of topic segmentation, bypassing the data sparsity issue of supervised 
models through the use of distant supervision from topic segmentation.

\indent\textbf{Topic Segmentation} aims to uncover the latent topical structure of a document by splitting it into a sequence of thematically coherent segments. Due to the shortage of annotated training data, early topic segmentation models are mostly unsupervised. 
As such, they merely exploit surface feature sets, like either the lexical overlap \cite{hearst-1997-text, eisenstein-barzilay-2008-bayesian} or topical signals contained in sentences \cite{riedl-biemann-2012-topictiling, du-etal-2013-topic} to measure sentence similarity and thus monitor topical and semantic changes throughout a document. More recently, the initial data sparsity issue has been lifted by large-scale corpora sampled from Wikipedia \cite{koshorek-etal-2018-text, arnold-etal-2019-sector}, containing well-structured articles with section marks as ground-truth segment boundaries. As a result, neural-based supervised topic segmenters \cite{koshorek-etal-2018-text, arnold-etal-2019-sector, xing-etal-2020-improving} have been heavily explored, due to their efficiency and robust performance. Following this trend, we choose a top-performing neural topic segmenter proposed by \citet{xing-etal-2020-improving}, directly addressing the context modeling problem commonly ignored in previous literature, as our base model for the discourse tree inference. 

\begin{figure}[t]
    \setlength{\belowcaptionskip}{-5pt}
    \centering
    \includegraphics[width=\linewidth]{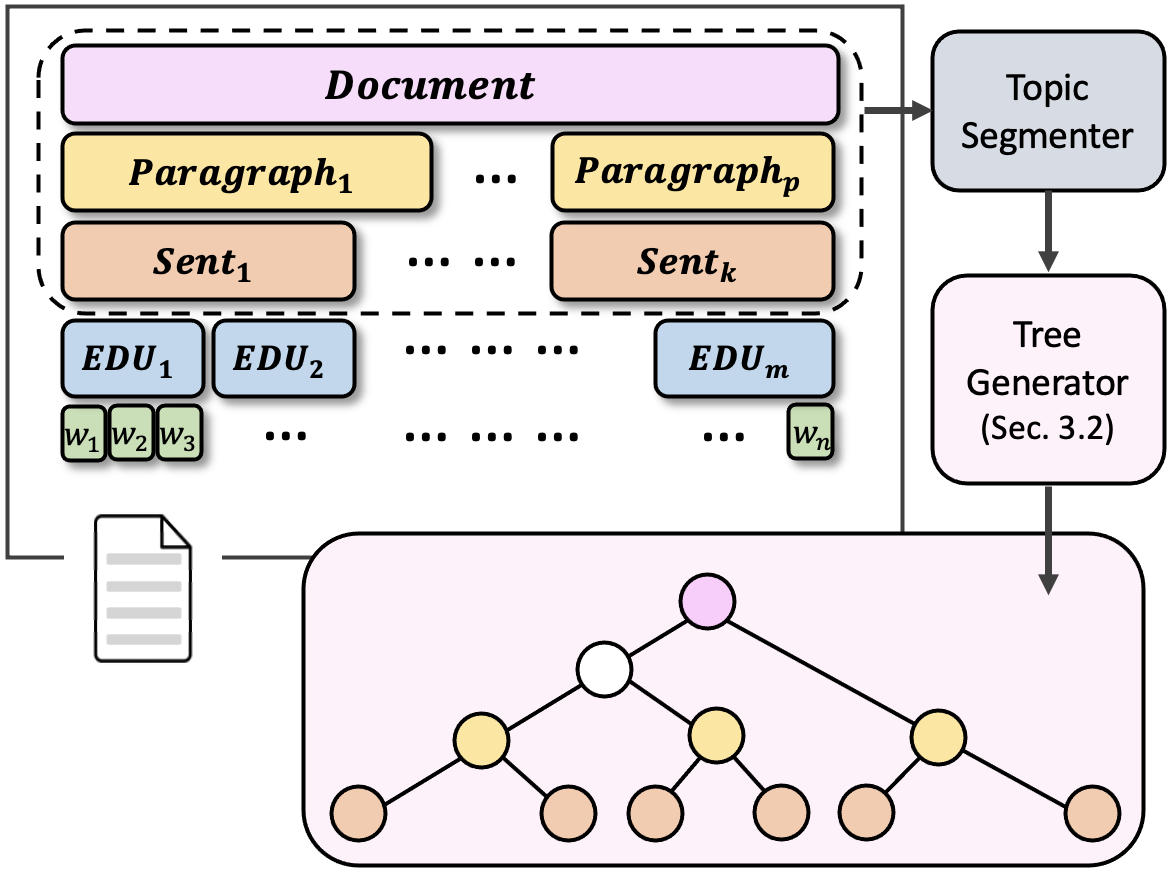}
    \caption{\label{fig:overview} Overview of our approach to generate high-level discourse structures on sentence-to-document level (S-D) from distant topic segmentation supervision.}
\end{figure}

\section{Methodology}
\label{method}
This section describes our new approach to generate high-level discourse structures for arbitrary documents using distant supervision from topic segmentation. An overview of our methodology is presented in Figure~\ref{fig:overview}. At the top, we visualize a naturally organized document, where words are aggregated into clauses (or EDUs), which are in turn concatenated into sentences. 
Sentences are then further combined into paragraphs\footnote{We overload the term ``paragraph'', also referring to what in other places is called ``sections''.}, forming the complete document. 
Here, we aggregate sentences as the atomic units to obtain high-level discourse trees (bottom of Figure~\ref{fig:overview}). 
We do not consider the intra-sentence task (EDU-to-sentence), as individually tackled by \citet{lin-etal-2019-unified} with great success.

\begin{figure}[t]
    \setlength{\belowcaptionskip}{-5pt}
    \centering
    \includegraphics[width=0.75\linewidth]{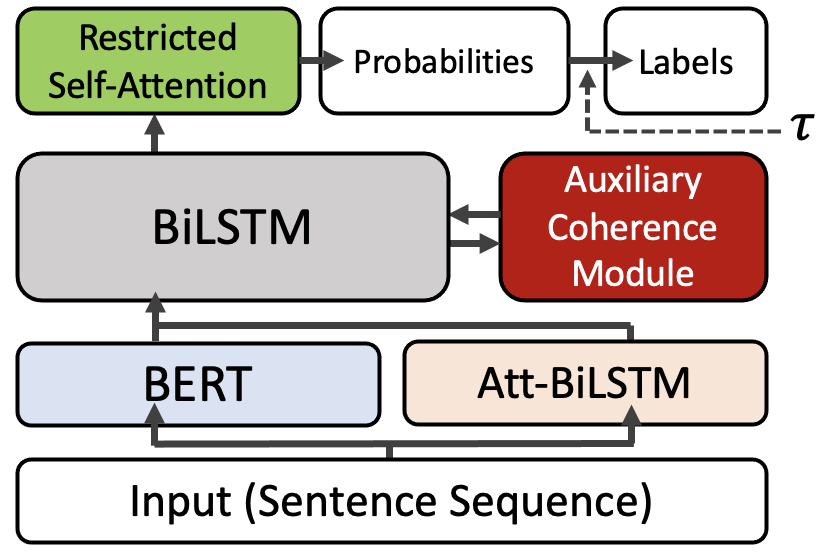}
    \caption{The high-level architecture of the neural topic segmentation model proposed in \citet{xing-etal-2020-improving}.}
    \label{fig:ts_model}
\end{figure}

\subsection{Topic Segmentation Model}
\label{topic_seg}
The topic segmentation task is commonly interpreted as a sequence labeling problem. Formally, given a document $d$ in the form of a sequence of sentences $\{s_1, s_2, ... , s_k\}$, a topic segmentation model assigns a probability score to each sentence $s_i$ for $i$ in $[1, ..., k-1]$, generating a sequence of probabilities $P = \{p(s_1), p(s_2), ... , p(s_{k-1})\}$ with $p(s_i) \in [0,1]$ indicating how likely sentence $s_i$ is the end of a segment\footnote{The last sentence $s_k$ does not need to be scored, since it is by definition the end of the last segment.}.

Based on the set of probability scores $P$ and an additional threshold $\tau$, determined on a held-out development set, topic segmentation models make binary predictions, labeling sentences with a probability larger than $\tau$ as the ``end of segment'', or ``within segment'' otherwise. Here, instead of converting the probabilities into discrete sentence boundary labels, we utilize the real-valued probability scores as our signal for distant supervision to infer RST-style discourse trees.

The high-level architecture of the neural topic segmentation model \cite{xing-etal-2020-improving} we use in this paper is shown in Figure~\ref{fig:ts_model}. Architecturally based on the proposal in \citet{koshorek-etal-2018-text}, the selected approach extends a hierarchical BiLSTM network by adding a BERT encoder as well as an auxiliary coherence prediction module and restricted self-attention, shown to improve the model's effectiveness for context modeling.
Since the topic segmentation task is generally applied on sentence level, any discourse tree generated based on its output can by definition only cover structures on and above sentence level (i.e., where leaf nodes represent sentences). 
The effectiveness of our selected topic segmentation model is demonstrated in Table~\ref{tab:ts_ds_performance}, showing our approach to yield superior performance over its direct competitors (described in Section~\ref{sec:baselines}) across all three evaluation corpora (described in Section~\ref{sec:datasets}). 

\begin{table}[t]
    \setlength{\belowcaptionskip}{-5pt}
    {\renewcommand{\arraystretch}{1.2}
    \centering
    \scalebox{0.84}{
    \begin{tabular}{lccc}
        \toprule
        \textbf{Model} & \textbf{Wiki} & \textbf{RST-DT} & \textbf{GUM} \\
        \midrule
        TextTiling \shortcite{hearst-1997-text} & 62.5 $\pm$ 0.00 & 44.3 $\pm$ 0.00 & 51.6 $\pm$ 0.00 \\
        BayesSeg \shortcite{eisenstein-barzilay-2008-bayesian} & 38.6 $\pm$ 0.00 & 37.5 $\pm$ 0.00 & 49.8 $\pm$ 0.00 \\
        GraphSeg \shortcite{glavas-etal-2016-unsupervised} & 43.2 $\pm$ 0.00 & 58.7 $\pm$ 0.00 & 53.9 $\pm$ 0.00 \\
        \citet{koshorek-etal-2018-text} & 27.9 $\pm$ 0.12 & 26.9 $\pm$ 0.49 & 48.0 $\pm$ 1.37 \\
        \citet{xing-etal-2020-improving} & \textbf{25.1 $\pm$ 0.07} & \textbf{25.4 $\pm$ 0.33} & \textbf{40.8 $\pm$ 1.64} \\
        \bottomrule
    \end{tabular}}}
    \caption{Average topic segmentation performance $\pm$ standard deviation over 5 runs, using the $P_k$ error score \cite{Beeferman1999} (lower values indicate better performance) on three evaluation corpora. Best performance per column in \textbf{bold}.}
    \label{tab:ts_ds_performance}
\end{table}

\subsection{Tree Generation}
\label{tree_agg}
\begin{figure}[t]
    \setlength{\belowcaptionskip}{-5pt}
    \centering
    \includegraphics[width=0.75\linewidth]{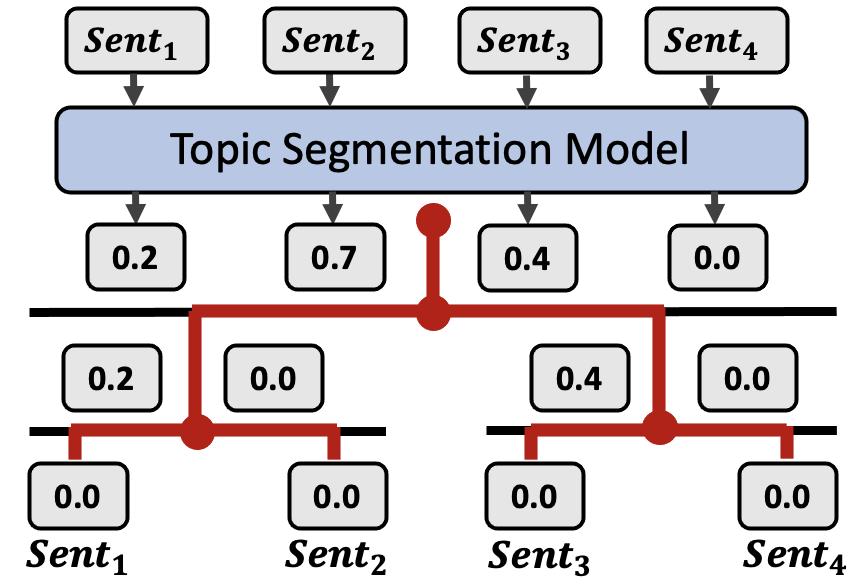}
    \caption{Greedy top-down tree generation approach.}
    \label{fig:tree_gen}
\end{figure}

To convert the sequential outputs of the topic segmentation model into tree structures, 
we follow the intuition that the value at position $i$ in the output of the topic segmentation model, namely the likelihood of $s_i$ being the end of a thematically-coherent segment, can be interpreted as the topical distance between $s_i$ and $s_{i+1}$.
A natural way to exploit the output of the topic segmentation model is to create a binary discourse tree by applying a greedy top-down algorithm where text spans (i.e., sub-trees) are determined recursively by splitting the two sentences with the largest topical distance. A small-scale example for this approach is shown in Figure~\ref{fig:tree_gen}. Here, we first search for the sentence $s_{max} = \argmax_{s_i \in d}{p(s_i)}$ with the maximum probability in $P$, making it our segmentation point. We then segment the sequence $P$ into two sub-sequences: $P_l = \{p(s_1), p(s_2),  ... , p(s_{max})\}$ (left portion) and $P_r = \{p(s_{max+1}), p(s_{max+2}), ... , p(s_{k})\}$ (right portion). Next, we mark $s_{max}$ as a previously selected segmentation point by setting $p(s_{max}) = 0.0$. We then recursively repeat this process for the two sub-sequences in a divide-and-conquer style, until all sentence probabilities are set to $0.0$. 
Noticeably, the bottom-up greedy strategy is equivalent to the top-down approach in our case.

\begin{figure}[t!]
    \setlength{\belowcaptionskip}{-5pt}
    \centering
    \includegraphics[width=\linewidth]{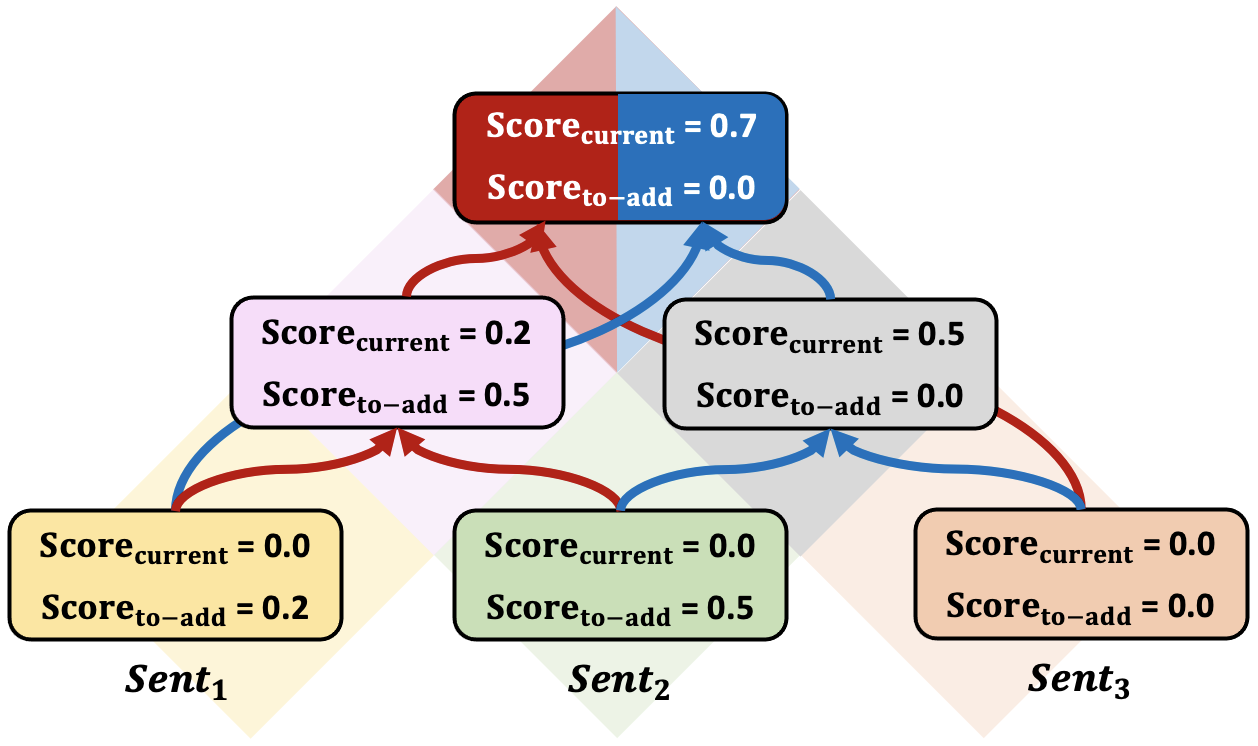}
    \caption{Problematic CKY aggregation without the use of discounting. All trees receive the same final score. 
    }
    \label{fig:cky}
\end{figure}

Besides the greedy approach described above, a commonly used tree-aggregation technique to convert a real-valued sequence into a binary tree structure is the optimal CKY dynamic programming algorithm, as previously used for distantly supervised discourse parsing in \citet{huber-carenini-2020-mega} and \citet{xiao2021predicting}. However, applying the CKY algorithm to the real-valued topic-break probabilities is problematic, since the output of any topic segmentation model only contains a single sequence. The intuitive way to fill the CKY matrix is to start merging any two consecutive units $s_i$ and $s_{i+1}$ with the score $p(s_i)$ of the former unit $s_i$ and assigning the new (merged) span the aggregation score of the latter, here $p(s_{i+1})$. Applying this strategy using the dynamic programming approach, every tree-candidate receives the same likelihood to represent the document at hand, due to the commutativity property of the subtree aggregation (see Figure \ref{fig:cky} for an example). 
In order to address this issue, additional hyper-parameters can be introduced, such as an attribute to quantify the benefit of merging low-probability sentences early on, and delaying likely topic-breaks. We explore this extended version of the CKY approach using a set of fixed discount factors. In preliminary experiments, we found that this adaption of the CKY approach, despite being theoretically superior, is in practice inferior to the greedy approach, showing unstable performance and resulting in heavily balanced/imbalanced trees for large/small discount factors, respectively. As a result, we leave the task of finding a more effective discounting function to future work and focus on the superior greedy top-down approach (as illustrated in Figure~\ref{fig:tree_gen}) in this paper. 

\section{Evaluation}
\subsection{Datasets} 
\label{sec:datasets}
We use three diverse datasets to train the topic segmentation models. We randomly sample a subset of Wikipedia articles from the Wikipedia dump\footnote{\url{https://dumps.wikimedia.org/enwiki/latest/enwiki-latest-pages-articles.xml.bz2}}, strictly following the sampling scheme in \citet{koshorek-etal-2018-text} (from here on called \textit{Wiki}). Our \textit{Wiki} corpus consists of $20,000$ documents, consistent in size with previously proposed topic segmentation training corpora, such as the \textit{Wiki-Section} dataset \cite{arnold-etal-2019-sector}, originally used in the top-performing model by \citet{xing-etal-2020-improving}. In contrast to the popular \textit{Wiki-Section} dataset, which extracts a strictly domain-limited subset of Wikipedia, exclusively covering webpages on \textit{cities} and \textit{diseases}, we lift this topic restriction by uniformly sampling from all of Wikipedia. We split our \textit{Wiki} dataset into training, validation and test sets using the default 80\%, 10\%, 10\% data-split. We further train the topic segmentation model on the \textit{RST-DT} \cite{carlson2002rst} and \textit{GUM} \cite{Zeldes2017} corpora. While \textit{RST-DT} exclusively contains news articles from the Wall Street Journal dataset, the additional \textit{GUM} treebank used in our experiments covers a mixture of 12 different genres, including interviews, news stories, academic writing and conversations. Since the two discourse corpora do not have any explicit human-annotated topic segment boundaries, we use paragraph breaks contained in the textual representation of the data as topic-shift indicators for the training of topic segmentation models. Please note that we do not use any human-annotated discourse structures during the training procedure of the topic segmentation model. 

The key statistics of all three datasets are presented in Table~\ref{tab:datasets}.

\subsection{Baselines}
\label{sec:baselines}
We compare our distantly supervised model against three sets of baselines: (1) Simple, but oftentimes competitive  structural and random baselines, including \textit{Right-Branching}, \textit{Left-Branching} and \textit{Random} Trees.
(2) Completely supervised models, including the traditional \textit{Two-Stage} model \cite{wang-etal-2017-two} trained within- and cross-domain, as well as the recently proposed neural \textit{SpanBERT} approach\footnote{We additionally show the results for two more supervised parsers proposed in \citet{kobayashi2020top} and \citet{jiang2021hierarchical} in Appendix \ref{app:results_parse}.} \cite{guz-carenini-2020-coreference}. (3) Distantly supervised models, exploiting signals from auxiliary tasks with available, large-scale training data. We compare our results against the model by \citet{huber-carenini-2020-mega}, training the \textit{Two-Stage} parser on the \textit{MEGA-DT} discourse treebank\footnote{\url{www.github.com/nlpat/MEGA-DT}} inferred from sentiment analysis, and the model by \citet{xiao2021predicting}, directly computing discourse structures from an extractive summarization model trained on the \textit{CNN-DM} dataset.

For topic segmentation, we compare the model proposed in \citet{xing-etal-2020-improving} against three widely applied unsupervised baselines: \textit{TextTiling} \cite{hearst-1997-text}, \textit{BayesSeg} \cite{eisenstein-barzilay-2008-bayesian} and \textit{GraphSeg} \cite{glavas-etal-2016-unsupervised}, and a competitive supervised model by \citet{koshorek-etal-2018-text} (see Table~\ref{tab:ts_ds_performance}).

\subsection{Experimental Design}
Given that our newly proposed RST-style tree generation strategy can only be applied on and above sentence level, we cannot directly compare our model with results provided in the literature, generally evaluating complete discourse trees from EDU-to-document level. 
In order to fairly compare against baselines and previously proposed models, and to provide a better understanding of how our proposal performs on different textual levels of the document (as depicted in Figures \ref{fig:intro_example} and \ref{fig:overview}), we design a set of comparisons extending the regular evaluation technique for complete discourse trees. 
More specifically, instead of just computing the micro-average performance for a complete document (i.e., EDU-to-document), we aim to investigate the structural performance on sentence-to-document (S-D), sentence-to-paragraph (S-P) and paragraph-to-document (P-D) level, allowing us to observe discourse tree generation trends on different levels of the document.
Therefore, complete (EDU-to-document) gold tree structures need to be trimmed to: (1) only contain nodes covering a complete unit of the lower-bound of measure (e.g., sentences in S-D/S-P) and (2) do not contain any nodes covering multiple units of the upper-bound of measure (e.g., paragraphs in S-P). Hence, we propose to trim and restrict discourse trees with the following two strategies: 

\noindent
\textbf{Trimming Trees to the Lower-Bound:}
Assuming we want to trim the discourse tree up to the sentence level (however, similarly applicable for paragraphs), we aim to remove all intra-sentence sub-trees that do not cover at least a complete sentence. In other words, assuming we have a sequence of EDUs: $E = \{e_1, e_2, ...e_m\}$, we keep a node $n$ spanning $e_i$ to $e_j$ in the discourse tree iff $sent(e_i) \neq sent(e_j)$ or $e_i.is\_beginning(sent_k) \land e_j.is\_end(sent_k)$, with $sent(\cdot)$ returning the sentence assignment of an EDU. Subsequently, we update the node $n = (e_i, e_j)$ to $n = (sent(e_i), sent(e_j))$. While this approach works for complete sentence
sub-trees, we follow the additional rules presented in \citet{sporleder-lascarides-2004-combining} for potentially leaky sentence-level trees (about $5\%$ of gold-standard trees in RST-DT \cite{joty-etal-2015-codra}). 

\noindent
\textbf{Leaky Sub-Tree Aggregation:}
Following the approach presented in \citet{sporleder-lascarides-2004-combining}, we replace leaky sentences with a single node $n = (sent(e_i), sent(e_j))$ and attach it depending on the largest intra-sentence subtree (e.g. if for a sentence containing 5 EDUs a sub-tree containing the first 3 EDUs is attached to the previous sentence and the sub-tree containing the last 2 EDUs is combined with the following sentence, we attach it to the previous sentence). In case of ties we attach the sentence to the right neighbour, as done in \citet{sporleder-lascarides-2004-combining}. Furthermore, since parts of the leaky sentences are propagated into larger sub-trees in the original structure, we recursively update the complete tree to be consistent in regards to the sentence assignment described above.

\noindent
\textbf{Restricting Trees to the Upper-Bound:}
Regarding the restriction of trees to selected upper-bounds (here exemplified on paragraph level), we remove any node $n$ covering sentences $s_i$ to $s_j$ iff $para(s_i) \neq para(s_j)$ with $para(\cdot)$ returning the paragraph assignment of a sentence.

\begin{table}[t]
    \centering
    \setlength{\belowcaptionskip}{-5pt}
    {\renewcommand{\arraystretch}{1.2}
    \scalebox{0.9}{
    \begin{tabular}{lccc}
        \toprule
        \textbf{Dataset} & \textbf{Wiki} & \textbf{RST-DT} & \textbf{GUM} \\
        \midrule
        \# of Docs. & 20,000 & 385 & 150 \\
        \# of Para./Doc. & 31.1 & 9.99 & 12.3 \\
        \# of Sents./Doc. & 144.9 & 22.5 & 49.3 \\
        \# of EDUs/Doc. & $\times$ & 56.6 & 114.2 \\
        \# of EDUs/Para. & $\times$ & 5.67 & 9.29 \\
        \# of EDUs/Sent. & $\times$ & 2.51 & 2.32 \\ 
        \bottomrule
    \end{tabular}}}
    \caption{Statistics of the three training/testing datasets.}
    \label{tab:datasets}
\end{table}

\subsection{Experiments and Results}
\label{results}
We show the RST parseval discourse structure performance of each model on the sentence-to-paragraph (S-P), paragraph-to-document (P-D) and finally sentence-to-document (S-D) level in Tables \ref{tab:rst_dt_results} and \ref{tab:gum_results} for RST-DT and GUM respectively\footnote{We present original parseval scores, as recommended in \citet{morey-etal-2017-much} in Appendix \ref{app:results_parse}.}. Our results are thereby subdivided into sub-tables according to the type of supervision used. The top sub-tables contain unsupervised baselines generating either random trees or completely right-/left-branching structures on the evaluated levels. The second set of results (in the center sub-tables) contain supervised discourse parsers, and the bottom sub-tables show distantly supervised models, including our results. To evaluate the ability of our proposal to generate domain-independent discourse structures, we compare the Topic Segmenter (TS) trained on in-domain data (TS\textsubscript{RST-DT/GUM}\footnote{Please note that even though we use the RST-DT/GUM datasets, we do not use any discourse tree annotation.}), the out-of-domain Wiki corpus (TS\textsubscript{Wiki}) as well as a ``fine-tuned'' approach, first trained on the Wiki corpus, and subsequently fine-tuned on RST-DT or GUM (TS\textsubscript{Wiki+RST-DT/GUM}). 
Finally, to further assess the role of the context modeling component (coherence module and restricted self-attention) in regards to the performance of topic segmentation as the distant supervised task for discourse tree generation, we ablate the context modeling component (red/green parts in Figure~\ref{fig:ts_model}) as ``Ablation -- TS\textsubscript{Wiki}" in the last row of both tables.

\begin{table}[t]
    \setlength{\belowcaptionskip}{-5pt}
    \centering
    {\renewcommand{\arraystretch}{1.2}
    \resizebox{\linewidth}{!}{
    \begin{tabular}{lccc}
        \toprule
        \textbf{Model} & \textbf{S-P} & \textbf{P-D} & \textbf{S-D} \\
        \midrule
        \multicolumn{4}{c}{Baselines}\\
        \midrule
        Random*  & \underline{77.11} & 63.90 & \underline{60.20}\\
        Right-Branching & 73.57 & \underline{65.50} & 59.46 \\
        Left-Branching & 72.41 & 64.07 & 58.07 \\
        \midrule
        \multicolumn{4}{c}{Supervised RST-style Parsers}\\
        \midrule
        Two-Stage\textsubscript{GUM} \shortcite{wang-etal-2017-two} & 88.82 & 65.63 & 69.58 \\
        Two-Stage\textsubscript{RST-DT} \shortcite{wang-etal-2017-two} &90.64 & 68.09 &  72.11 \\
        SpanBERT\textsubscript{RST-DT} \shortcite{guz-carenini-2020-coreference} & \underline{\textbf{90.75}} & \underline{\textbf{76.03}} & \underline{\textbf{77.19}} \\
        \midrule
        \multicolumn{4}{c}{Distantly Supervised RST-style Parsers}\\
        \midrule
        \citet{xiao2021predicting}\textsubscript{CNN/DM} & 74.23 & 66.15 & 59.10 \\
        Two-Stage\textsubscript{MEGA-DT} \shortcite{huber-carenini-2020-mega} & \underline{85.00} & 65.50 & 66.99  \\
        TS\textsubscript{RST-DT} & 84.34 & 62.52 & 65.96 \\
        TS\textsubscript{Wiki} & 83.43 & \underline{69.78}  & \underline{68.13} \\
        TS\textsubscript{Wiki+RST-DT} & 83.84 & 66.54 & 65.84 \\
        Ablation -- TS\textsubscript{Wiki} & 83.51 & 68.61 & 67.47 \\
        \bottomrule
    \end{tabular}}}
    \caption{Evaluation results using the RST Parseval micro-average precision measure on the RST-DT dataset.
    Subscripts indicate training dataset. TS = Topic Segmentation Model. * = Average performance over 10 runs. Best performance per sub-table \underline{underlined}, best performance per column \textbf{bold}.}
    \label{tab:rst_dt_results}
\end{table}

Not surprisingly, when evaluating discourse structures on the RST-DT dataset (Table \ref{tab:rst_dt_results}), supervised models generally outperform unsupervised baselines and distantly supervised models. Assessing the bottom sub-table in more detail, it becomes clear that while the Two-Stage parser trained on MEGA-DT achieves the best performance on the sentence-to-paragraph level, our model solely trained on the Wiki dataset performs best on the paragraph-to-document and sentence-to-document level, showing that (1) obtaining discourse structures from topic segmentation effectively supports high-level discourse parsing and (2) general out-of-domain training on large-scale data (TS\textsubscript{Wiki}) performs better than models trained or fine-tuned on in-domain data (TS\textsubscript{RST-DT} and TS\textsubscript{Wiki+RST-DT} respectively). We believe that a possible explanation for the surprising under-performance of the in-domain training described above could originate from the limited size of the RST-DT dataset, as well as the mismatch between the “granularity” of segment information in Wikipedia and RST-DT (see Table 2), where the average number of sentences per segment is about twice as large in Wikipedia than RST-DT. Plausibly, the larger segments in Wikipedia better support higher-level discourse structures (which cover larger text spans), leading to superior performance, with the RST-DT fine-tuning step (at finer granularity) introducing the mismatch, and therefore not delivering the expected benefits.

\begin{table}[t]
    \centering
    {\renewcommand{\arraystretch}{1.2}
    \resizebox{\linewidth}{!}{
    \begin{tabular}{lccc}
        \toprule
        \textbf{Model} & \textbf{S-P} & \textbf{P-D} & \textbf{S-D} \\
        \midrule
        \multicolumn{4}{c}{Baselines}\\
        \midrule
        Random* & \underline{67.53} & 60.96 & 57.99 \\
        Right-Branching & 64.15 & \underline{72.71} & \underline{59.39} \\
        Left-Branching & 62.07 & 54.35 & 51.56 \\
        \midrule
        \multicolumn{4}{c}{Supervised RST-style Parsers}\\
        \midrule
        Two-Stage\textsubscript{RST-DT} \shortcite{wang-etal-2017-two} & 74.20 & 63.29 & 63.65\\
        Two-Stage\textsubscript{GUM} \shortcite{wang-etal-2017-two} & \underline{76.70} & \underline{\textbf{72.94}} & \underline{\textbf{68.38}}\\
        \midrule
        \multicolumn{4}{c}{Distantly Supervised RST-style Parsers}\\
        \midrule
        \citet{xiao2021predicting}\textsubscript{CNN-DM} & 67.89 & 57.80 & 53.82 \\
        Two-Stage\textsubscript{MEGA-DT} \shortcite{huber-carenini-2020-mega} & 73.37 & \underline{69.88} & 64.69 \\
        TS\textsubscript{GUM} & 72.54 & 67.60 & 62.79 \\
        TS\textsubscript{Wiki} & \underline{\textbf{76.98}} & 63.53 & \underline{65.84} \\
        TS\textsubscript{Wiki+GUM} & 74.48 & 67.29 & 64.69 \\
        Ablation -- TS\textsubscript{Wiki} & 75.94 & 64.71 & 65.38 \\
        \bottomrule
    \end{tabular}}}
    \caption{Evaluation results using the RST Parseval micro-average precision measure on the GUM dataset. 
    Subscripts indicate training dataset. TS = Topic Segmentation Model. 
    * = Average performance over 10 runs. Best performance per sub-table \underline{underlined}, best performance per column \textbf{bold}.}
    \label{tab:gum_results}
\end{table}

Our evaluation results on the GUM dataset (Table~\ref{tab:gum_results}) show similar trends to the evaluation on RST-DT. 
Our in-domain trained and further fine-tuned models in the bottom sub-table do not achieve improved performances compared to the out-of-domain TS\textsubscript{Wiki} model. We believe a possible explanation for this phenomenon is the mix of domains within the small GUM training portion, resulting in the fine-tuning step mixing noisy signals from different genres, hence not providing consistent improvements over Wikipedia. Interestingly however, there are some observations which  differ from RST-DT:
(1) Unlike for RST-DT, our distantly supervised model trained on \textit{Wiki} even outperforms supervised approaches on sentence-to-document level. (2) Right-branching trees outperform random trees, which hasn't been the case on the RST-DT dataset and (3) the paragraph-to-document (P-D) level differs from previous results, with the right-branching baseline reaching a performance close to the best supervised model (Two-Stage\textsubscript{GUM}), outperforming the out-of-domain supervised model (Two-Stage\textsubscript{RST-DT}) and all distantly supervised approaches.

Regarding the ablation studies, our results shown in the last row of Tables \ref{tab:rst_dt_results} and \ref{tab:gum_results} imply that the context modeling component, shown to boost the topic segmentation performance, can also consistently benefit the high-level discourse structure inference on S-D level. 

We refer readers to Appendix~\ref{app:results_parse} for results on additional evaluation levels (e.g., EDU-to-sentence and EDU-to-paragraph) using the RST and original parseval scores.

\begin{table}[t]
    \centering
    \setlength{\tabcolsep}{0.3em}
    {\renewcommand{\arraystretch}{1.2}
    \resizebox{\linewidth}{!}{
    \begin{tabular}{lccccc}
    \toprule
        \textbf{Genre} & \textbf{RB} & \textbf{2S\textsubscript{GUM}} & \textbf{2S\textsubscript{RST-DT}} &  \textbf{TS\textsubscript{Wiki}} & \textbf{TS\textsubscript{GUM}} \\
        \midrule
        Travel guides & \textbf{78.1} & 75.0 & 65.6 & 53.1 & 68.8 \\
        Biographies & 75.0 & \textbf{78.6} & 60.7 & \textbf{78.6} & 71.4 \\
        Fiction & \textbf{80.6} & \textbf{80.6} & 61.1 & 61.1 & 61.1 \\
        How-to guides & 69.4 & 64.3 & 59.2 & 66.3 & \textbf{75.5} \\
        Academic writing & 70.4 & \textbf{81.5} & 74.1 & 70.4 & 63.0 \\
        News stories & 57.4 & 57.4 & 61.8 & 63.2 & \textbf{69.1} \\
        Political speeches & 80.0 & \textbf{85.0} & 70.0 & 60.0 & 55.0 \\
        Textbooks & \textbf{78.6} & 71.4 & 64.3 & 57.1 & 71.4 \\
        Interviews & 78.8 & \textbf{83.3} & 66.7 & 60.6 & 60.6 \\
        \bottomrule
    \end{tabular}}}
    \caption{RST Parseval micro-average precision on paragraph-to-document level for genres in the GUM corpus. Sample(s) in ``Vlog'' \& ``Conversation'' only contain a single paragraph and are therefore omitted. RB=Right-Branching, 2S=Two-Stage parser, TS=Topic Segmentation model.}
    \label{tab:details_gum}
\end{table}

\begin{figure}[t]
    \setlength{\belowcaptionskip}{-5pt}
    \centering
    \includegraphics[width=.6\linewidth]{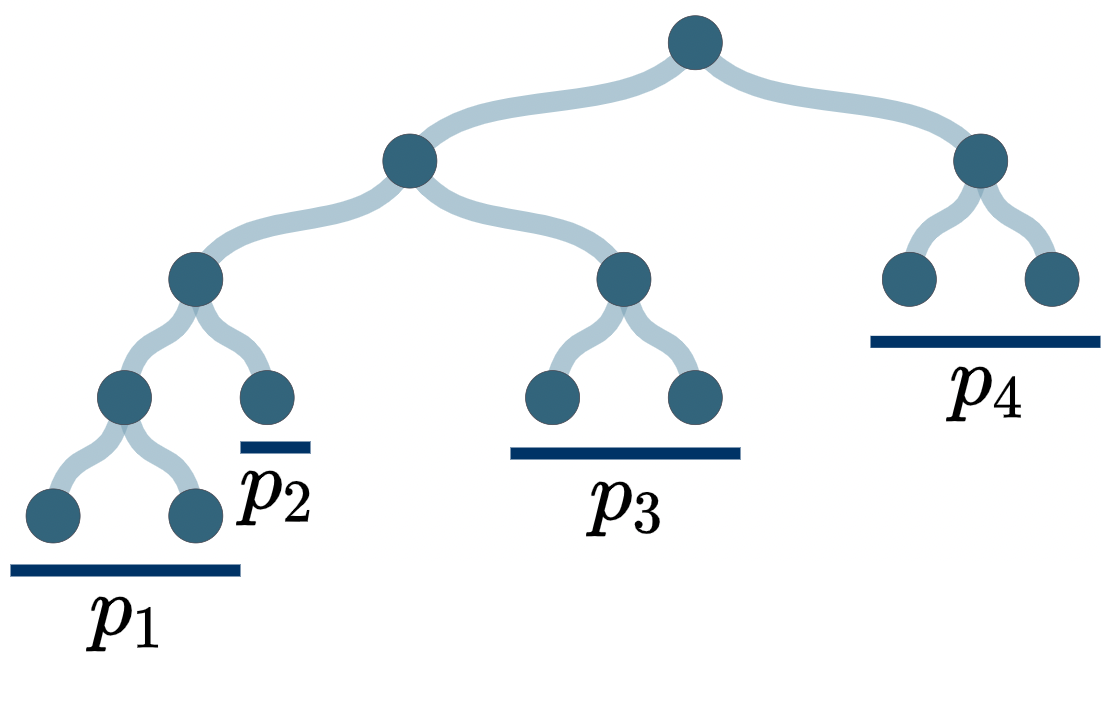}
    \caption{Positive example (wsj\_1346) with 100\% structural overlap between prediction and RST-DT gold-label annotation. Gold paragraphs $p_n$ are indicated by solid lines. Leaves are sentences.}
    \label{fig:pos_example_qualitative}
\end{figure}

\begin{figure}[t]
    \setlength{\belowcaptionskip}{-5pt}
    \centering
    \includegraphics[width=0.75\linewidth]{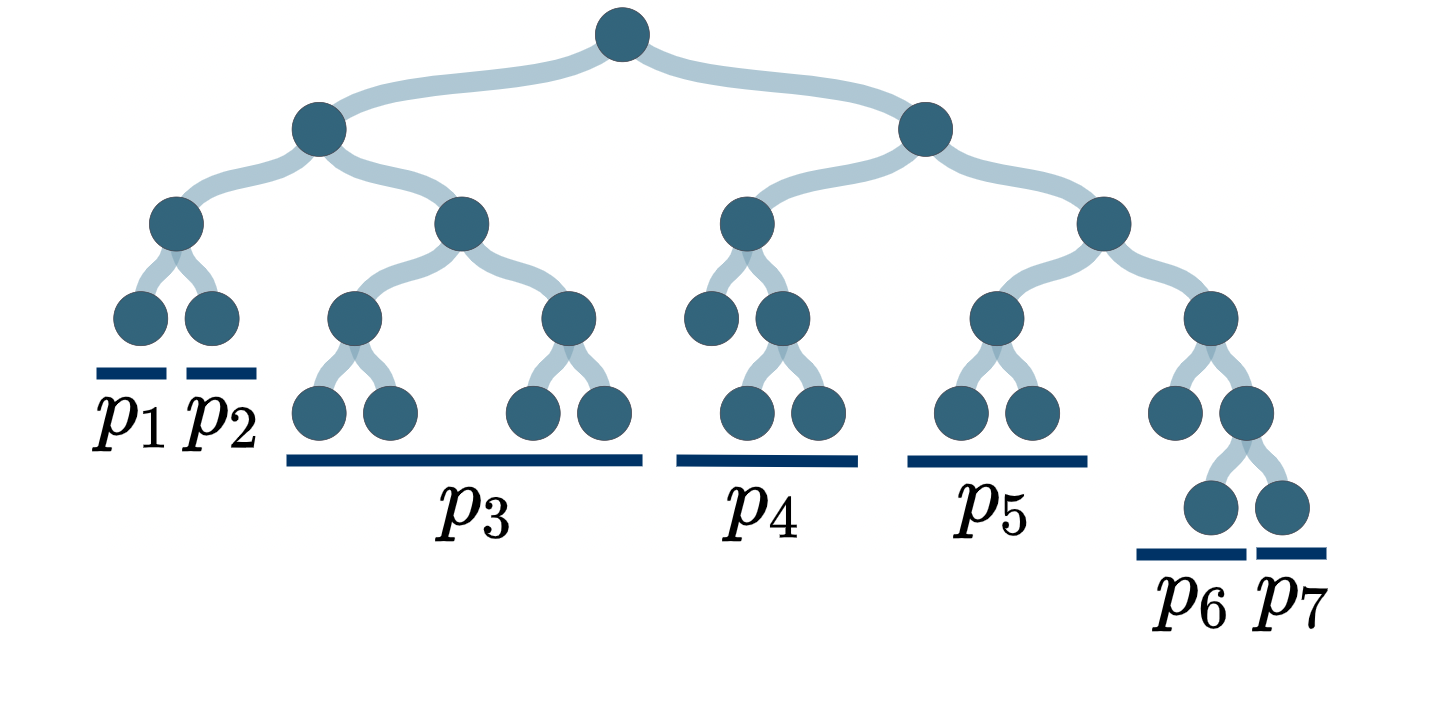}
    \includegraphics[width=0.75\linewidth]{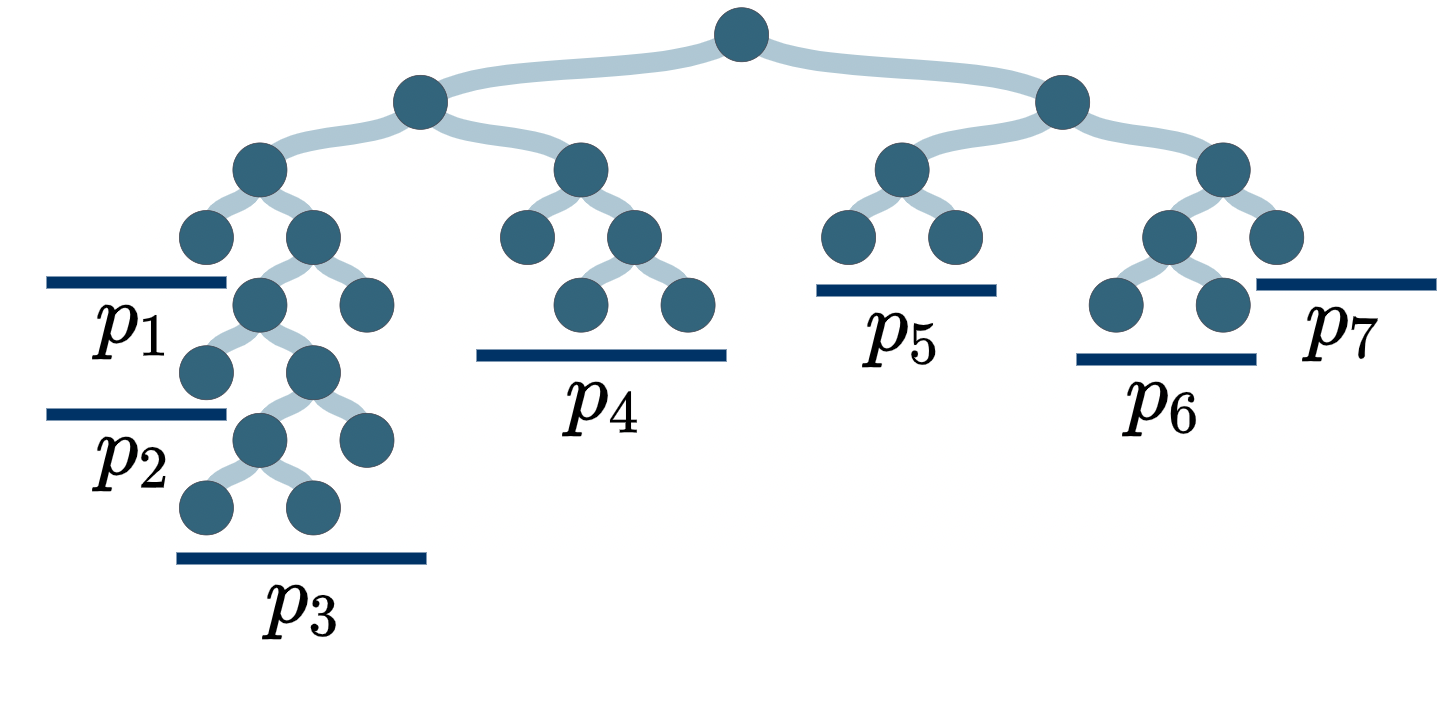}
    \caption{Random example (wsj\_1380) with 80.77\% structural overlap between prediction (top) and RST-DT gold-label annotation (bottom). Gold paragraphs $p_n$ are indicated by solid lines. Leaves are sentences.}
    \label{fig:rand_example_qualitative}
\end{figure}

\begin{figure}[t]
    \setlength{\belowcaptionskip}{-5pt}
    \centering
    \includegraphics[width=0.7\linewidth]{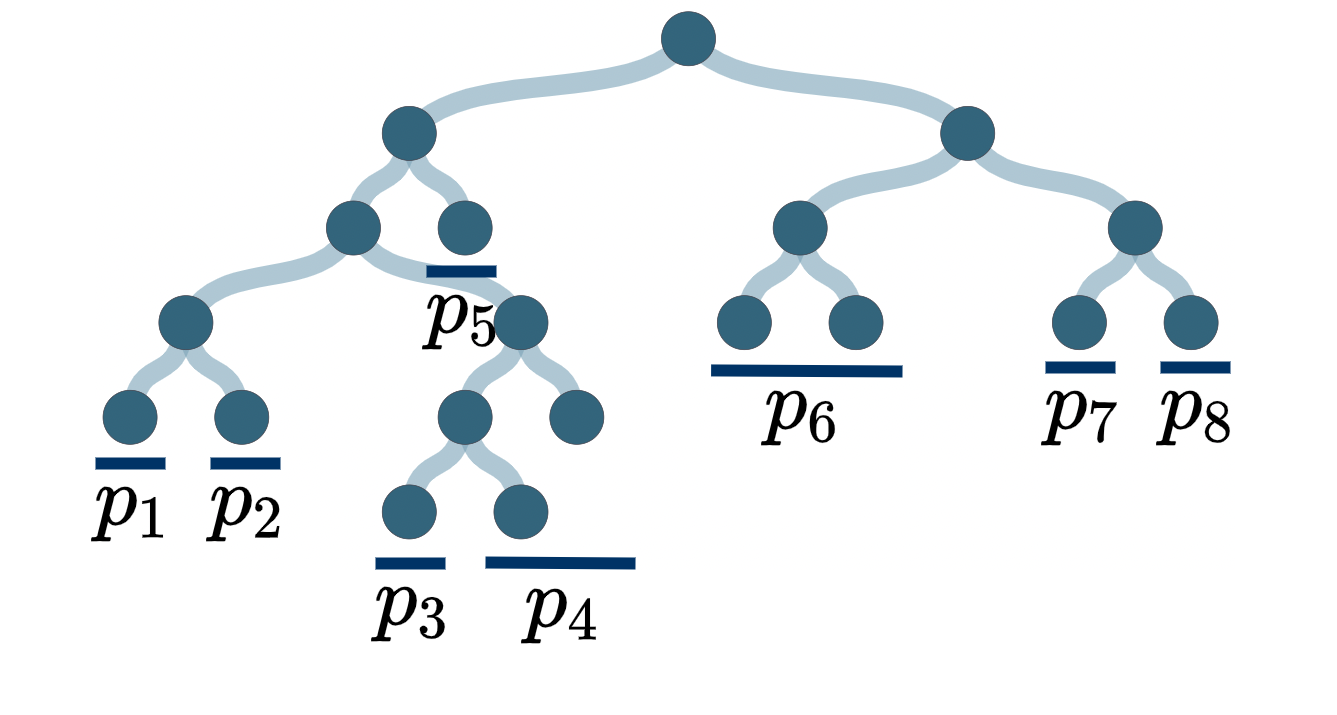}
    \includegraphics[width=0.7\linewidth]{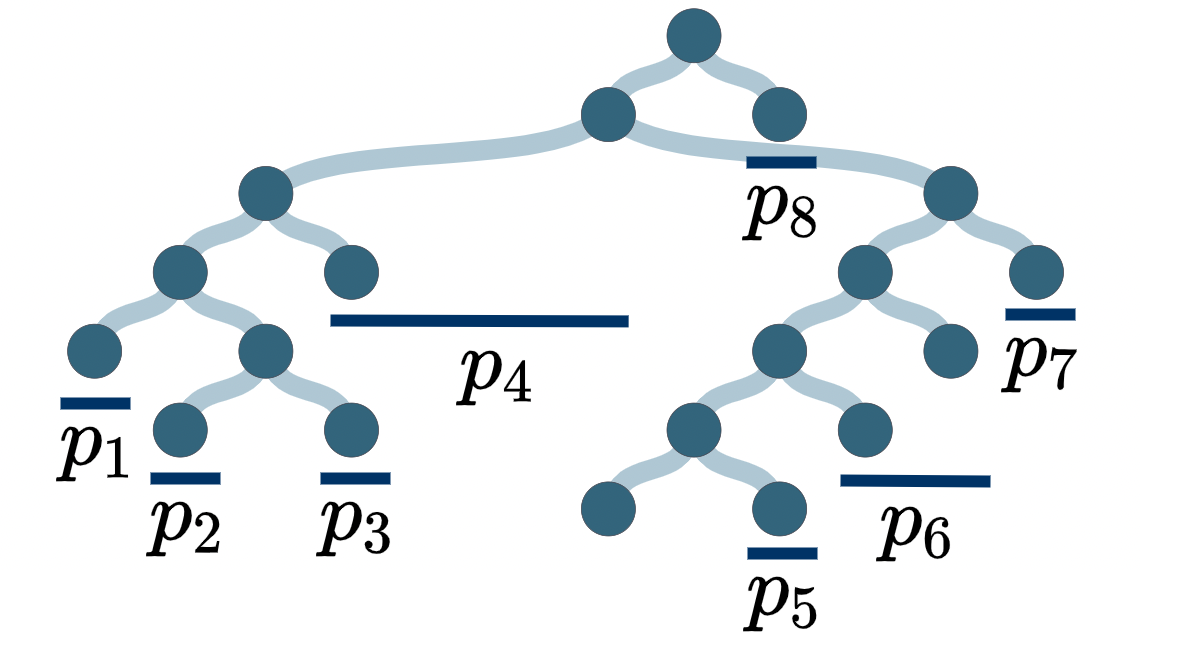}
    \caption{Negative example (wsj\_1365) with 55.56\% structural overlap between prediction (top) and RST-DT gold-label annotation (bottom). Gold paragraphs $p_n$ are indicated by solid lines. Leaves are sentences.}
    \label{fig:neg_example_qualitative}
\end{figure}

To further investigate the performance on paragraph-to-document level for the GUM corpus, we show a comparison by genre for the surprisingly high performing right-branching baseline, the two supervised models and our methods based on \textit{Wiki} and \textit{GUM} in Table \ref{tab:details_gum}. Right-branching trees thereby achieve the best performance in 3 out of the 9 genres, including textbooks and fiction. Supervised methods perform best on 5 out of the 9 genres, including highly structured domains such as academic writing and interviews. Our distantly supervised model trained on Wikipedia reaches the best performance on biographies and the topic segmentation model trained on GUM achieves the highest score on how-to guides and news articles. Furthermore, as expected, the supervised parser trained on RST-DT (i.e., the news domain) performs well on the news genre. Overall, while these mixed results appear to align well with our intuition on the prevalent structures in certain genres, further research is required to better understand the relationship between modelling decisions and their impact on different discourse structures across genres. 

To complement our quantitative evaluation, we show a set of predicted trees and their respective gold-label structures for well captured above-sentence discourse (Figure \ref{fig:pos_example_qualitative}), randomly selected documents (Figure \ref{fig:rand_example_qualitative}) and poorly captured samples (Figure \ref{fig:neg_example_qualitative}). Similar results for the GUM corpus are shown in Figures \ref{fig:pos_example_qualitative_gum}, \ref{fig:rand_example_qualitative_gum} and \ref{fig:neg_example_qualitative_gum} in Appendix \ref{app:trees}.

Overall, inspecting a large number of tree-structures along with their gold-labels, we recognize that the generated trees are slightly more balanced than the respective gold-label trees (see for example Figures \ref{fig:rand_example_qualitative} and \ref{fig:neg_example_qualitative}), however generally represent non-trivial structures, oftentimes well-aligned with major topic shifts and high-level discourse.

\section{Conclusions and Future Work}
In this paper, we show that topic segmentation can provide useful signals for high-level discourse constituency tree structure generation. Comparing multiple aggregation approaches, our proposal using a greedy top-down algorithm performs well when applied on two popular gold-standard discourse treebanks, namely RST-DT and GUM. We provide a detailed evaluation based on textual levels in documents, giving insights into the strength and weaknesses of simple baselines, previously proposed models and our new, distantly supervised approach using topic segmentation, on sentence, paragraph and document level. We  show additional insights into our modelling approach through an ablation study, per-genre evaluations and qualitative tree generation samples.

For the future, in the short term we plan to investigate alternative non-greedy tree aggregation strategies, such as variations of the CKY approach mentioned in section \ref{tree_agg}. Next, we want to explore if the synergy between discourse parsing and topic segmentation is bidirectional, incorporating discourse signals to improve topic segmentation models. Finally, inspired by the approach described in \citet{kobayashi-etal-2019-split}, we also plan to use dense representations of neural topic segmentation models to infer discourse structures with nuclearity and relation labels, to obtain more complete trees by solely exploiting distant supervision from topic segmentation.

\section*{Acknowledgements}
We thank the anonymous reviewers and the UBC NLP group for their insightful comments and suggestions. This research was supported by the Language \& Speech Innovation Lab of Cloud BU, Huawei Technologies Co., Ltd and the Natural Sciences and Engineering Research Council of Canada (NSERC). Nous remercions le Conseil de recherches en sciences naturelles et en génie du Canada (CRSNG) de son soutien.

\bibliography{aaai22, anthology}

\clearpage

\onecolumn

\appendix

\section{Complete Quantitative Results}
\label{app:results_parse}
We show the complete quantitative results, including the previously shown Sentence-to-Paragraph (S-P), Paragraph-to-Document (P-D) and Sentence-to-Document (S-D) performance, as well as the additional EDU-to-Sentence (E-S), EDU-to-Paragraph (E-P), EDU-to-Document (E-D) and EDU-to-Sentence-to-Document (E-S-D) sub-tree results here. Tables \ref{tab:rst_dt_results_full_rst_parse} and \ref{tab:gum_results_full_rst_parse} contain the RST parseval scores on the RST-DT and GUM corpus, respectively. Table \ref{tab:rst_dt_results_full_orig_parse} contains the full results of the original parseval score applied to RST-DT and Table \ref{tab:gum_results_full_orig_parse} shows the original parseval performance on GUM. Besides the additional performance evaluations also covering low-level structures for the baseline models (E-S, E-P, E-D and E-S-D), we show an additional model, combining the out-of-the-box discourse segmenter by \citet{wang-etal-2018-toward} with our above-sentence tree structures generated from topic-segmentation in the last row of Tables \ref{tab:rst_dt_results_full_rst_parse}, \ref{tab:gum_results_full_rst_parse}, \ref{tab:rst_dt_results_full_orig_parse} and \ref{tab:gum_results_full_orig_parse} (\textit{Combined -- TS\textsubscript{Wiki}/Disc-Seg}). Since we combine strictly intra-sentence structures from the discourse segmenter with strictly above-sentence trees, this approach does not support leaky EDUs, and hence constitutes E-S-D structures (as compared to non sentence limited E-D trees).

\begin{table*}[ht]
    \centering
    \begin{tabular}{lccccccc}
        \toprule
        \textbf{Model} & \textbf{E-S} & \textbf{S-P} & \textbf{P-D} & \textbf{E-P} & \textbf{S-D} & \textbf{E-D} & \textbf{E-S-D} \\
        \midrule
        \multicolumn{8}{c}{Baselines}\\
        \midrule
        Random* & \underline{70.08} & \underline{77.11} & 63.90 & \underline{64.05} & \underline{60.20} & \underline{58.03} & 70.10\\
        Right-Branching & 64.12 & 73.57 & \underline{65.50} & 58.41 & 59.46 & 54.64 & \underline{74.37}\\
        Left-Branching & 63.85 & 72.41 & 64.07 & 57.73 & 58.07 & 53.73 & 70.58\\
        \midrule
        \multicolumn{8}{c}{Supervised RST-style Parsers}\\
        \midrule
        Top-Down\textsubscript{RST-DT} \shortcite{kobayashi2020top} & $\times$ & $\times$ & $\times$ & $\times$ & $\times$ & 86.10 & \underline{\textbf{86.40}} \\
        Two-Stage\textsubscript{GUM} \shortcite{wang-etal-2017-two} & 92.57 & 88.82 & 65.63 & 88.62 & 69.58 & 81.24 & $\times$ \\
        Two-Stage\textsubscript{RST-DT} \shortcite{wang-etal-2017-two} & 94.48 & 90.64 & 68.09 & 91.06 & 72.11 & 83.90 & $\times$ \\
        SpanBERT\textsubscript{RST-DT} \shortcite{guz-carenini-2020-coreference} & \underline{\textbf{96.27}} & \underline{\textbf{90.75}} & \underline{\textbf{76.03}} & \underline{\textbf{92.87}} & \underline{\textbf{77.19}} & \underline{\textbf{87.69}} & $\times$ \\
        \midrule
        \multicolumn{8}{c}{Distantly Supervised RST-style Parsers}\\
        \midrule
        \citet{xiao2021predicting}\textsubscript{CNN/DM} & 85.95 & 74.23 & 66.15 & 78.19 & 59.10 & 60.68 & 75.13\\
        Two-Stage\textsubscript{MEGA-DT} \shortcite{huber-carenini-2020-mega} & \underline{89.57} & \underline{85.00} & 65.50 & \underline{84.73} & 66.99 & \underline{77.90} & $\times$ \\
        TS\textsubscript{RST-DT} & $\times$ & 84.34 & 62.52 & $\times$ & 65.96 & $\times$ & $\times$ \\
        TS\textsubscript{Wiki} &  $\times$ & 83.43 & \underline{69.78} & $\times$ & \underline{68.13} & $\times$ & $\times$ \\
        TS\textsubscript{Wiki+RST-DT} &  $\times$ & 83.84 & 66.54 & $\times$ & 65.84 & $\times$ & $\times$\\
        Ablation -- TS\textsubscript{Wiki} &  $\times$ & 83.51 & 68.61 & $\times$ & 67.47 & $\times$ & $\times$ \\
        Combined -- TS\textsubscript{Wiki}/Disc-Seg & 87.10 & 83.43 & 69.78 & 81.67 & 68.13 & $\times$ & \underline{75.56} \\
        \bottomrule
    \end{tabular}
    \caption{Evaluation results using the \textbf{RST Parseval} micro-average precision measure on the RST-DT dataset. \\
    Subscripts indicate training dataset. TS=Topic Segmentation Model. Disc-Seg=Discourse Segmentation Model by \citet{wang-etal-2018-toward}. 
    $\times$=Not feasible combination. *=Average performance over 10 runs. Best performance per sub-table \underline{underlined}, best performance per column \textbf{bold}}
    \label{tab:rst_dt_results_full_rst_parse}
\end{table*}

\begin{table*}[ht]
    \centering
    \begin{tabular}{lccccccc}
        \toprule
        \textbf{Model} & \textbf{E-S} & \textbf{S-P} & \textbf{P-D} & \textbf{E-P} & \textbf{S-D} & \textbf{E-D} & \textbf{E-S-D} \\
        \midrule
        \multicolumn{8}{c}{Baselines}\\
        \midrule
        Random* & \underline{72.16} & \underline{67.53} & 60.96 & \underline{61.60} & 57.99 & \underline{57.24} & 69.93 \\
        Right-Branching & 66.30 & 64.15 & \underline{72.71} & 56.23 & \underline{59.39} & 54.71 & \underline{71.58} \\
        Left-Branching & 65.96 & 62.07 & 54.35 & 55.08 & 51.56 & 50.77 & 64.57 \\
        \midrule
        \multicolumn{8}{c}{Supervised RST-style Parsers}\\
        \midrule
        Two-Stage\textsubscript{RST-DT} \shortcite{wang-etal-2017-two} & \underline{\textbf{93.51}} & 74.20 & 63.29 & 82.74 & 63.65 & 77.16 & $\times$ \\
        Two-Stage\textsubscript{GUM} \shortcite{wang-etal-2017-two} & 93.25 & \underline{76.70} & \underline{\textbf{72.94}} & \underline{\textbf{83.49}} & \underline{\textbf{68.38}} & \underline{\textbf{79.04}} & $\times$ \\
        \midrule
        \multicolumn{8}{c}{Distantly Supervised RST-style Parsers}\\
        \midrule
        \citet{xiao2021predicting}\textsubscript{CNN-DM} & 88.09 & 67.89 & 57.80 & 76.52 & 53.82 & 59.28 & 72.39 \\
        Two-Stage\textsubscript{MEGA-DT} \shortcite{huber-carenini-2020-mega} & \underline{88.78} & 73.37 & \underline{69.88} & 78.35 & 64.69 & \underline{73.89} & $\times$ \\
        TS\textsubscript{GUM} & $\times$ & 72.54 & 67.60 & $\times$ & 62.79 & $\times$ & $\times$ \\
        TS\textsubscript{Wiki} & $\times$ & \underline{\textbf{76.98}} & 63.53 & $\times$ & \underline{65.84} & $\times$ & $\times$ \\
        TS\textsubscript{Wiki+GUM} & $\times$ & 74.48 & 67.29 & $\times$ & 64.69 & $\times$ & $\times$ \\
        Ablation -- TS\textsubscript{Wiki} & $\times$ & 75.94 & 64.71 & $\times$ & 65.38 & $\times$ & $\times$ \\
        Combined -- TS\textsubscript{Wiki}/Disc-Seg & 88.58 & 76.98 & 63.53 & 79.30 & 65.84 & $\times$ & \underline{\textbf{73.94}} \\
        \bottomrule
    \end{tabular}
    \caption{Evaluation results using the \textbf{RST Parseval} micro-average precision measure on the GUM dataset. \\
     Subscripts indicate training dataset. TS=Topic Segmentation Model. Disc-Seg=Discourse Segmentation Model by \citet{wang-etal-2018-toward}. $\times$=Not feasible combination. *=Average performance over 10 runs. Best performance per sub-table \underline{underlined}, best performance per column \textbf{bold}}
    \label{tab:gum_results_full_rst_parse}
\end{table*}

\begin{table*}[ht]
    \centering
    \begin{tabular}{lccccccc}
        \toprule
        \textbf{Model} & \textbf{E-S} & \textbf{S-P} & \textbf{P-D} & \textbf{E-P} & \textbf{S-D} & \textbf{E-D} & \textbf{E-S-D} \\
        \midrule
        \multicolumn{8}{c}{Baselines}\\
        \midrule
        Random* & \underline{19.35} & \underline{22.55} & 30.61 & \underline{17.13} & \underline{21.83} & \underline{16.06} & 41.07 \\
        Right-Branching & 3.25 & 7.27 & \underline{31.62} & 4.12 & 19.21 & 9.27 & \underline{48.74} \\
        Left-Branching & 2.53 & 3.20 & 28.79 & 2.56 & 16.45 & 7.45 & 41.16 \\
        \midrule
        \multicolumn{8}{c}{Supervised RST-style Parsers}\\
        \midrule
        MDParser -- TS\textsubscript{RST-DT} \shortcite{jiang2021hierarchical} & $\times$ & $\times$ & 40.52 & $\times$ & $\times$ & $\times$ & $\times$ \\
        Two-Stage\textsubscript{GUM} \shortcite{wang-etal-2017-two} & 79.97 & 60.76 & 31.88 & 73.78 & 39.38 & 62.48 & $\times$ \\
        Two-Stage\textsubscript{RST-DT} \shortcite{wang-etal-2017-two} & 85.10 & 67.44 & 36.76 & 79.40 & 44.54 & 70.97 & $\times$ \\
        SpanBERT\textsubscript{RST-DT} \shortcite{guz-carenini-2020-coreference} & \underline{\textbf{90.16}} & \underline{\textbf{68.69}} & \underline{\textbf{52.75}} & \underline{\textbf{83.83}} & \underline{\textbf{54.37}} & \underline{\textbf{75.39}} & $\times$ \\
        \midrule
        \multicolumn{8}{c}{Distantly Supervised RST-style Parsers}\\
        \midrule
        \citet{xiao2021predicting}\textsubscript{CNN-DM} & 62.11 & 9.59 & 32.09 & 49.72 & 18.49 & 21.36 & 49.21 \\
        Two-Stage\textsubscript{MEGA-DT} \shortcite{huber-carenini-2020-mega} & \underline{71.87} & \underline{47.38} & 31.62 & \underline{64.81} & 34.21 & 55.81 & $\times$ \\
        TS\textsubscript{RST-DT} & $\times$ & 45.06 & 28.02 & $\times$ & 32.17 & $\times$ & $\times$ \\
        TS\textsubscript{Wiki} & $\times$ & 41.86 & \underline{41.90} & $\times$ & \underline{36.49} & $\times$ & $\times$ \\
        TS\textsubscript{Wiki+RST-DT} & $\times$ & 43.31 & 34.96 & $\times$ & 31.93 & $\times$ & $\times$ \\
        Ablation -- TS\textsubscript{Wiki} & $\times$ & 42.15 & 39.07 & $\times$ & 35.17 & $\times$ & $\times$ \\
        Combined -- TS\textsubscript{Wiki}/Disc-Seg & 65.22 & 41.86 & 41.90 & 57.74 & 36.49 & $\times$ & \underline{\textbf{50.95}} \\
        \bottomrule
    \end{tabular}
    \caption{Evaluation results using the \textbf{original Parseval} micro-average precision measure on the RST-DT dataset. \\
    Subscripts indicate training dataset. TS=Topic Segmentation Model. Disc-Seg=Discourse Segmentation Model by \citet{wang-etal-2018-toward}. 
    $\times$=Not feasible combination. *=Average performance over 10 runs. Best performance per sub-table \underline{underlined}, best performance per column \textbf{bold}}
    \label{tab:rst_dt_results_full_orig_parse}
\end{table*}

\begin{table*}[ht]
    \centering
    \begin{tabular}{lccccccc}
        \toprule
        \textbf{Model} & \textbf{E-S} & \textbf{S-P} & \textbf{P-D} & \textbf{E-P} & \textbf{S-D} & \textbf{E-D} & \textbf{E-S-D} \\
        \midrule
        \multicolumn{8}{c}{Baselines}\\
        \midrule
        Random* & \underline{18.45} & \underline{17.83} & 25.46 & \underline{15.25} & 16.71 & \underline{14.49} & 43.11 \\
        Right-Branching & 1.28 & 7.35 & \underline{46.30} & 3.39 & \underline{18.78} & 9.42 & \underline{43.15} \\
        Left-Branching & 0.30 & 1.97 & 10.19 & 0.86 & 3.11 & 1.54 & 29.15 \\
        \midrule
        \multicolumn{8}{c}{Supervised RST-style Parsers}\\
        \midrule
        Two-Stage\textsubscript{RST-DT} \shortcite{wang-etal-2017-two} & \underline{\textbf{81.00}} & 33.69 & 27.78 & 61.90 & 27.53 & 54.33 & $\times$ \\
        Two-Stage\textsubscript{GUM} \shortcite{wang-etal-2017-two} & 80.22 & \underline{39.78} & \underline{\textbf{46.76}} & \underline{\textbf{63.56}} & \underline{\textbf{36.75}} & \underline{\textbf{58.08}} & $\times$ \\
        \midrule
        \multicolumn{8}{c}{Distantly Supervised RST-style Parsers}\\
        \midrule
        \citet{xiao2021predicting}\textsubscript{CNN-DM} & 65.72 & 7.60 & 16.75 & 46.69 & 7.64 & 18.56 & 45.97 \\
        Two-Stage\textsubscript{MEGA-DT} \shortcite{huber-carenini-2020-mega} & \underline{67.13} & 31.36 & \underline{40.74} & 52.22 & 29.49 & 47.79 & $\times$ \\
        TS\textsubscript{GUM} & $\times$ & 29.03 & 37.50 & $\times$ & 25.58 & $\times$ & $\times$ \\
        TS\textsubscript{Wiki} & $\times$ & \underline{\textbf{40.50}} & 31.02 & $\times$ & \underline{31.68} & $\times$ & $\times$ \\
        TS\textsubscript{Wiki+GUM} & $\times$ & 34.05 & 37.04 & $\times$ & 29.38 & $\times$ & $\times$ \\
        Ablation -- TS\textsubscript{Wiki} & $\times$ & 37.81 & 33.33 & $\times$ & 30.76 & $\times$ & $\times$ \\
        Combined -- TS\textsubscript{Wiki}/Disc-Seg & 66.54 & 40.50 & 31.02 & 54.32 & 31.68 & $\times$ & \underline{\textbf{47.81}} \\
        \bottomrule
    \end{tabular}
    \caption{Evaluation results using the \textbf{original Parseval} micro-average precision measure on the GUM dataset. \\
    Subscripts indicate training dataset. TS=Topic Segmentation Model. Disc-Seg=Discourse Segmentation Model by \citet{wang-etal-2018-toward}. 
    $\times$=Not feasible combination. *=Average performance over 10 runs. Best performance per sub-table \underline{underlined}, best performance per column \textbf{bold}}
    \label{tab:gum_results_full_orig_parse}
\end{table*}

\clearpage
\newpage

\section{Qualitative Analysis on GUM}
\label{app:trees}

\begin{figure*}[ht]
    \centering
    \includegraphics[width=.48\linewidth]{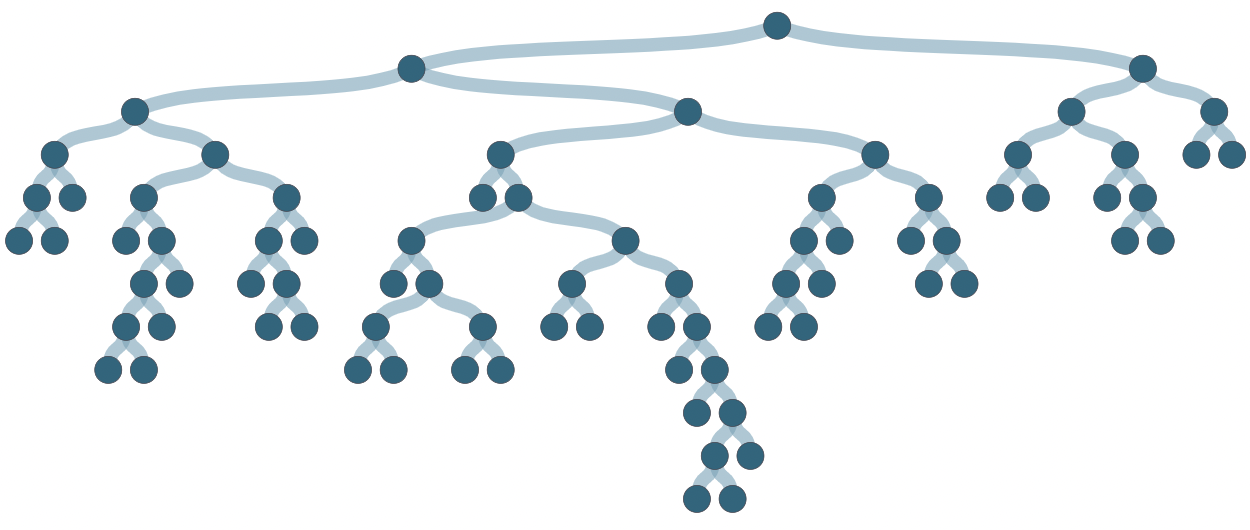}
    \includegraphics[width=.48\linewidth]{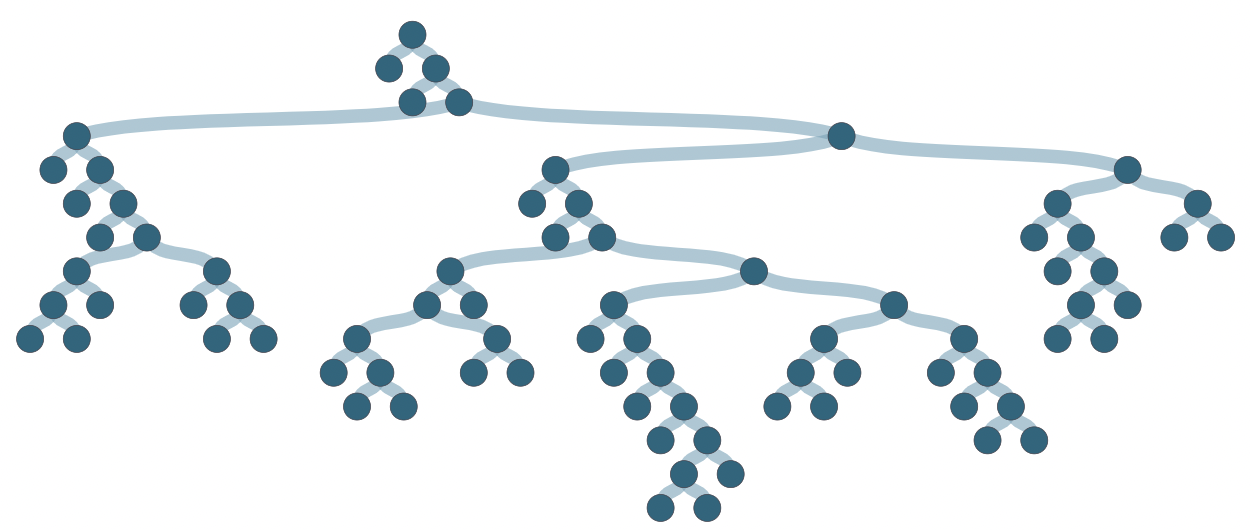}
    \caption{Positive example (GUM\_bio\_jesperser) with 76.92\% structural overlap between prediction (left) and GUM gold-label annotation (right).}
    \label{fig:pos_example_qualitative_gum}
\end{figure*}

\begin{figure*}[ht]
    \centering
    \includegraphics[width=.48\linewidth]{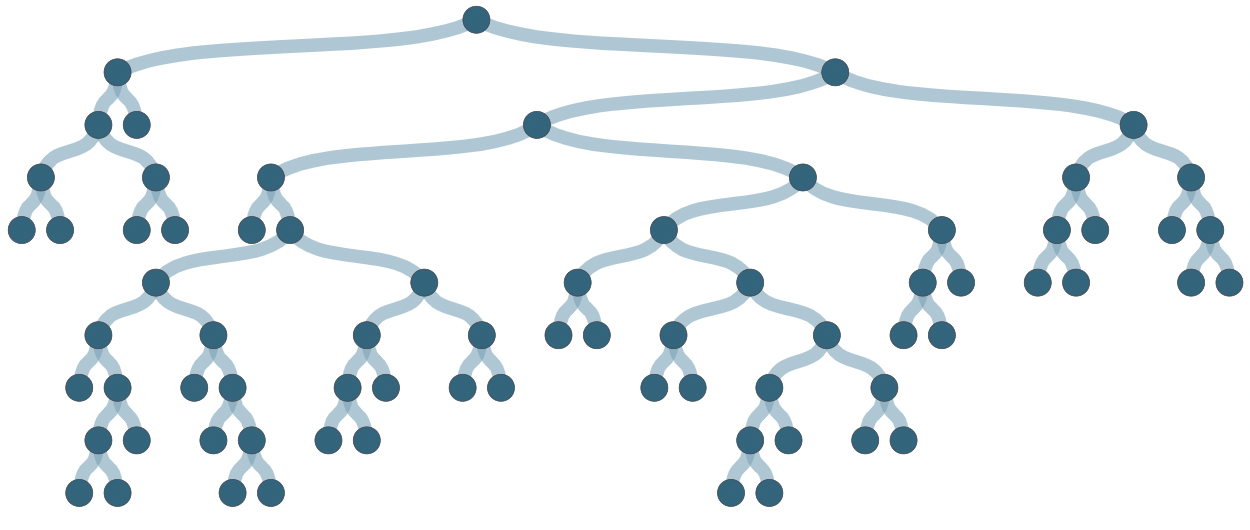}
    \includegraphics[width=.48\linewidth]{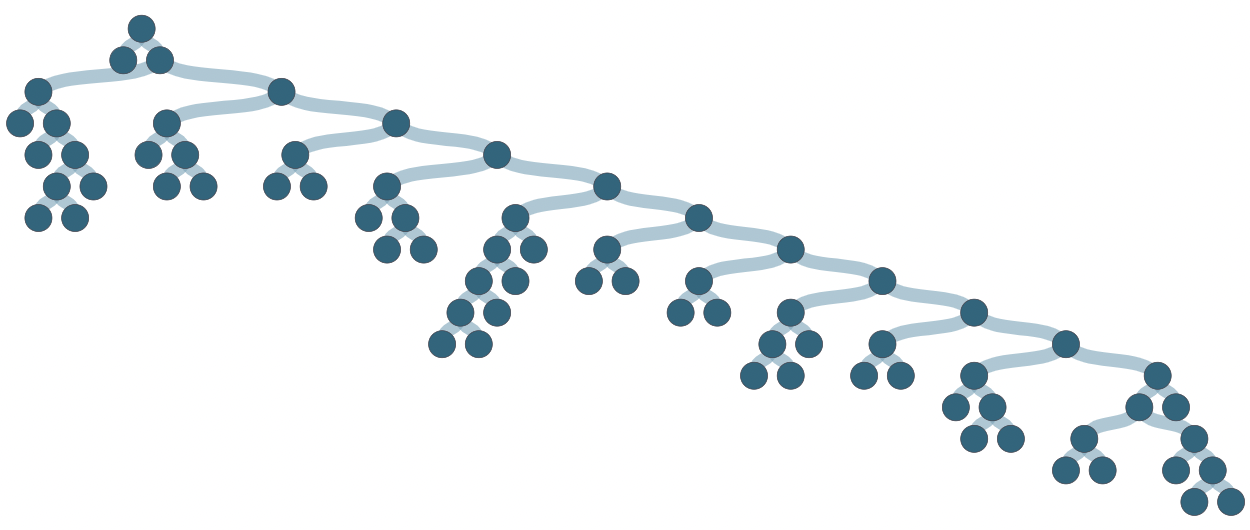}
    \caption{Random example (GUM\_voyage\_oakland) with 70.83\% structural overlap between prediction (left) and GUM gold-label annotation (right).}
    \label{fig:rand_example_qualitative_gum}
\end{figure*}

\begin{figure*}[ht]
    \centering
    \includegraphics[width=.48\linewidth]{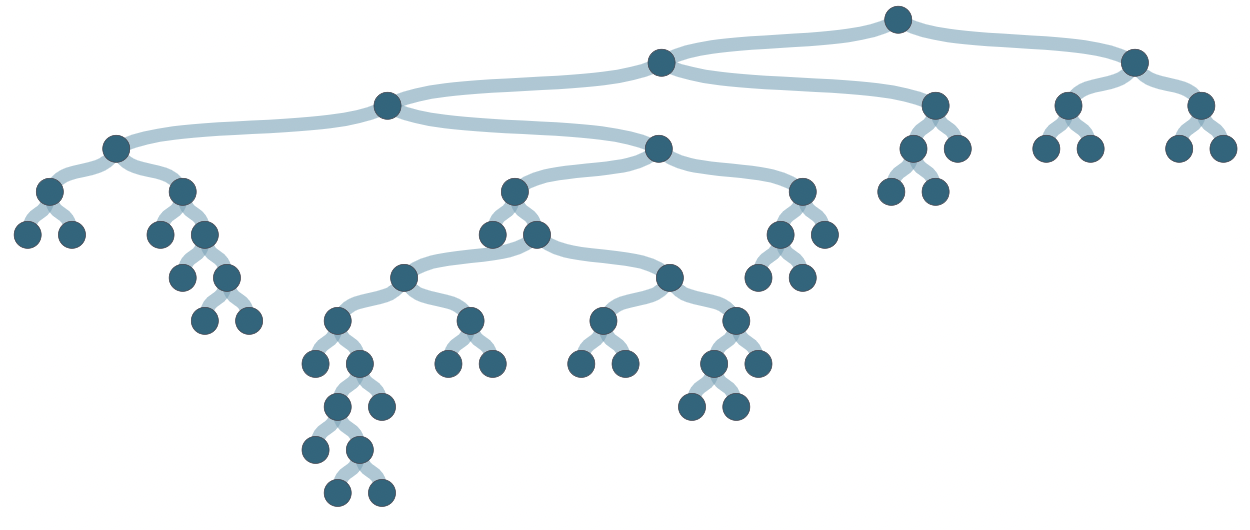}
    \includegraphics[width=.48\linewidth]{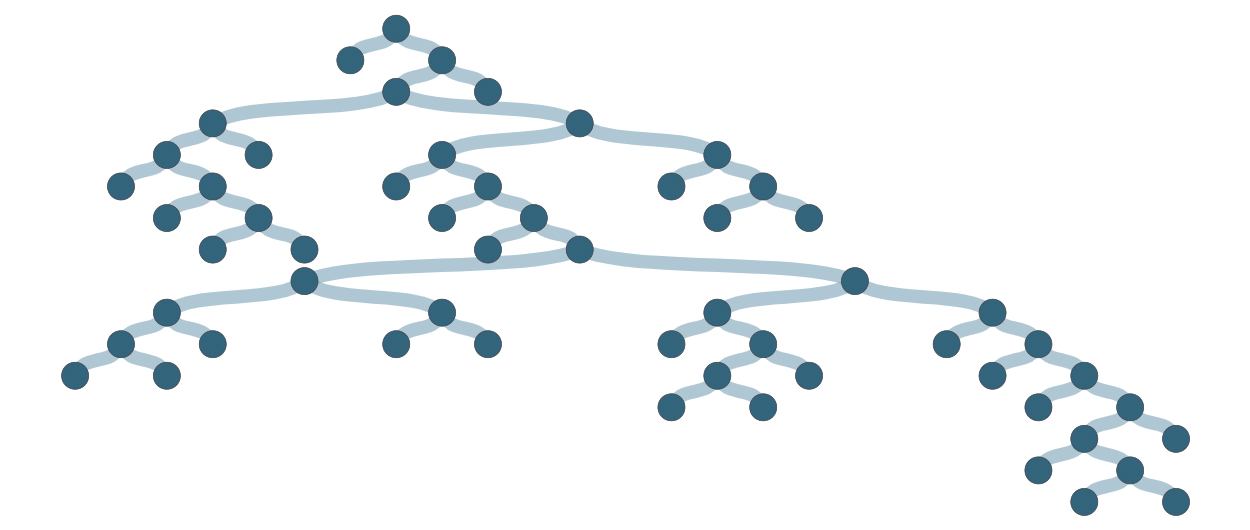}
    \caption{Negative example (GUM\_bio\_dvorak) with 57.14\% structural overlap between prediction (left) and GUM gold-label annotation (right).}
    \label{fig:neg_example_qualitative_gum}
\end{figure*}

\end{document}